\documentclass[letterpaper, 10 pt, conference]{ieeeconf}  

\IEEEoverridecommandlockouts                              

\usepackage{amsmath}
\usepackage{amssymb}
\usepackage{amsfonts}
\usepackage{algorithm}
\usepackage{algorithmic}
\usepackage{overpic}
\usepackage{mathtools}
\usepackage{array}

\usepackage{multirow}
\usepackage{wrapfig}
\usepackage{capt-of}
\let\labelindent\relax
\usepackage{enumitem}

\usepackage{scalerel}[2016/12/29]
\newcommand{\vTheta}{{\boldsymbol \Theta}}
\newcommand{\vtheta}{{\boldsymbol \theta}}
\newcommand{\vPhi}{{\boldsymbol{\Phi}}}
\newcommand{\vphi}{{\boldsymbol{\phi}}}

\newcommand{\W}{\scaleobj{0.85}{\mathbf{W}}}
\newcommand{\w}{{\mathbf{w}}}

\newcommand{\vbeta}{{\boldsymbol \beta}}

\newcommand\smallpercent{\vcenter{\hbox{\scalebox{0.8}{$\%$}}}}

\usepackage{xcolor}

\newcommand{\rev}[1]{{{#1}}}

\title{\LARGE \bf
Efficient Multi-Task and Transfer Reinforcement Learning with Parameter-Compositional Framework}

\author{Lingfeng Sun$^{*\dagger{1}}~\thanks{*Equal contribution. $\dagger$Work done while interning at Horizon Robotics.}$ \quad Haichao Zhang$^{*{2}}$ \quad  Wei Xu$^{2}$ \quad Masayoshi Tomizuka$^{1}$
\thanks{$^{1}$UC Berkeley
        {\tt\footnotesize \{lingfengsun, tomizuka\}@berkeley.edu}}%
\thanks{$^{2}$Horizon Robotics
        {\tt\footnotesize \{haichao.zhang, wei.xu\}@horizon.ai}}%
}

\begin{document}

\maketitle
\thispagestyle{empty}
\pagestyle{empty}

\begin{abstract}
        In this work, we investigate the potential of improving multi-task training and also leveraging it for transferring in the reinforcement learning setting. We identify several challenges towards this goal and propose a transferring approach with a parameter-compositional formulation. We investigate ways to improve the training of multi-task reinforcement learning which serves as the foundation for transferring. Then we conduct a number of transferring experiments on various manipulation tasks.
        Experimental results demonstrate that the proposed approach can have improved performance in the multi-task training stage, and further show effective transferring in terms of both sample efficiency and performance.
\end{abstract}


\section{Introduction}\label{sec:intro}

How to equip robots with various skills that are necessary for solving a diverse set of tasks is a long-time quest in the robotics field.
Using reinforcement learning to empower robots with various skills is an active and valuable research direction.
However, many robotics-related tasks, such as different manipulation skills, are commonly treated as isolated tasks and used individually for learning~\cite{fetch_env, gallouedec2021pandagym}.
While this setup is suitable for some particular research problems,
learning one policy for each skill and learning from scratch every time is not practical for real-world robots, as it does not scale well with the number of skills to be acquired and introduces additional \rev{wear and tear} if a physical robot is involved. This motivates the pursuit of more effective learning algorithms in research.



Reusing previously acquired knowledge in learning new tasks is attractive both in terms of learning efficiency and performance.
A natural approach is to transfer the previously learned policy for the learning of a new skill.
This is challenging as the transfer performance between tasks is strongly affected by relationships between individual tasks.
A more effective approach for acquiring various skills is to solve the multi-task reinforcement learning (MTRL) problem~\cite{metaworld}.
Recent works compare multi-task learning and gradient-based meta learning~\cite{bridge_mtl_meta,ANIL} under the supervised learning setting and conclude that for both methods, the learned representations are the key to transfer.

Inspired by the long-term quest on transferring learning in RL and encouraging progress on supervised multi-task learning and transferring~\cite{bridge_mtl_meta}, in this work, we aim to further investigate the potential of leveraging multi-task learning for transfer in the RL setting.
There are some algorithmic necessities for bringing the benefits of MTRL to the transfer learning on new tasks:
\begin{enumerate}
  \item \textbf{Performant MTRL Method.} The MTRL framework's high performance on a diverse set of manipulation tasks is a prerequisite. If MTRL fails to learn a well-performing policy, it
  is natural to expect that the transferring performance built upon it will be limited.
  \item \textbf{Proper Architecture.} Not all MTRL frameworks are capable of being used for transferring to new tasks. Some task-specific information contained in the inputs of policy (\emph{e.g.}, task-specific information such as task id) cannot be directly transferred to another new task where that piece of information is different.
 \end{enumerate}

\begin{figure}[t]
	\centering
	\begin{overpic}[width=8.5cm]{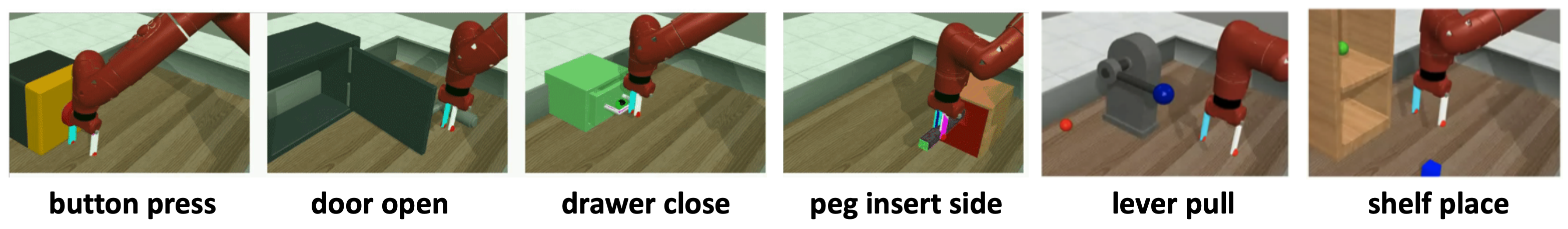}
	\end{overpic}
	\vspace{-0.1in}
	\caption{Some example tasks from Meta-World benchmark~\cite{metaworld}.}
	\vspace{-0.1in}
	\label{fig:example_tasks}
\end{figure}

In addition to algorithmic design, the performance of transfer is also strongly affected by the level of difficulty of the new task itself as well as its relation to the trained tasks.
Few-shot meta reinforcement learning for new tasks typically works under a task distribution of limited variations~\cite{metaworld}, \emph{e.g.} pushing the object to different goal locations.
In practice, tasks typically come from a family with a more significant variation, \emph{e.g.}, different types of manipulations skills as shown in Figure~\ref{fig:example_tasks}, posing a great challenge to these approaches both in terms of performance as well as learning efficiency, calling for the development of new approaches.

To overcome these challenges, one possible approach is to first design an MTRL training framework that succeeds in training high-performance policies for various tasks, with a transferable policy structure, and then design an efficient transfer algorithm that can benefit from the multi-task policies.
For this purpose, we build on top of a recent parameter-compositional MTRL framework  called PaCo~\cite{paco}.
PaCo maintains a policy subspace for all the individual task policies and by learning on multiple tasks with a compositional structure for task-adaptive parameter sharing.

Training multiple tasks with task-adaptive parameter sharing has two major benefits for learning skills: \textit{\textbf{i)}} The MTRL training process benefits from similarities between tasks and is usually more efficient in both the number of parameters and the roll-out samples required to solve multiple tasks~\cite{soft_module, paco}; \textit{\textbf{ii)}} \rev{The task-adaptive parameter sharing schemes learned for solving different tasks gives us a posterior relation between tasks} and can benefit both the multi-task training as well as transferring to unseen but similar tasks.
The PaCo approach fits well in our context both because of its good MTRL performance and architecture design regarding the necessities discussed above.
Firstly, it reaches state-of-the-art performance on the Meta-World \rev{MT10}~\cite{metaworld} consists of 10 different manipulation tasks, each with multiple goals.
Secondly, there is a clear separation of task-specific and task-agnostic parameters in the policy~\cite{paco}; thus the task-agnostic part of PaCo policies can be naturally transferred for learning new tasks.
The compositional vectors can be used to compute task relationships in policy subspace, as shown in Figure \ref{fig:paco_pca}.

Motivated by these signs of progress, in this work, we focus on designing  approaches for PaCo-based multi-task and transfer RL and investigating its potential.
Our main contribution in this work can be summarized as follows:
\begin{enumerate}
    \item \textbf{Improved MTRL}: further improve MTRL performance based on PaCo by incorporating a mechanism that adjusts task distributions during training;
    \item \textbf{Transfer RL}: investigate the possibility and potential of MTRL for transferring, which can be viewed as
further expanding recent results on multi-task learning for transferring from supervised setting~\cite{bridge_mtl_meta,ANIL} to reinforcement learning;
    \item \textbf{Empirical Validations}: validate the improved performance of transfer learning using the MTRL policy trained with the improved MTRL approach.
\end{enumerate}

\section{Related Work}
\label{sec:revisit}

\subsection{Multi-Task and Transfer Reinforcement Learning}

{\flushleft \textbf{Multi-Task RL.}} The goal of MTRL is to train a single policy $\pi_{\theta}(a|s)$ that can be applied to a set of different tasks.
Every single task can be defined by a unique MDP, with differences in either state/action space, transition, or reward functions.
In MTRL, we solve a set of MDPs from a task family using a universal policy $\pi_\theta(a|s, w)$, with $w$ the task-relevant context information.
By doing this, we can potentially have at least two benefits:
\textbf{\emph{i)}} better performance: by leveraging mutual connections between tasks, we can potentially improve the sample efficiency or final performance, \emph{e.g.} through parameter sharing.
\textbf{\emph{ii)}} transferring: the \rev{learned MTRL policy could serve} as a starting point to be transferred to a new task.
While the first point has been extensively explored in MTRL literature~\cite{metaworld, policy_sketch, pcgrad, share_knowledge_mtrl, Distral, soft_module, care, MT_OPT}, the second one is less investigated~\cite{actor_mimic, modular_rl, never_stop_learning},
potentially due to many practical challenges in MTRL itself, \emph{e.g.},  conflicts between tasks~\cite{pcgrad} and training stability~\cite{care, paco}, limiting its effectiveness on transferring.

{\flushleft \textbf{Transfer RL.}}
Transferring learning is a general approach to re-use previously learned knowledge in learning new situations.
It is a powerful paradigm in supervised learning, leading to many successful approaches such as pre-training + fine-tuning that is widely used in computer vision~\cite{finetune_cv, imagenet_transfer} and natural language processing~\cite{bert, gpt1}.
However, for reinforcement learning, simply fine-tuning a previously learned policy is not very useful~\cite{RL_finetune, simple_approach}. The reason is that different from supervised learning, RL requires sufficient exploration for successful learning.
And a policy transferred from a previously trained task could be overly deterministic for exploring in the current task.
There are also some previous attempts on leveraging MTRL for transferring,  with focuses on goal-conditional/contextual policy~\cite{fetch_env, rl_transfer, zeroshot_task_generalization,SF_transfer, usf, her}, policy modularity~\cite{modular_rl, policy_sketch} or reward learning~\cite{mtrl_reward}.
For example, a large portion of works leverages the context-conditional form for generalization $\pi_{\theta}(a|s, w)$ where $w$ denotes the task-relevant context information~\cite{zeroshot_task_generalization}. This approach relies on the
the generalization ability of the network \emph{w.r.t.} the input context and sufficient coverage of it during training for transferring to new tasks (new contexts).

\begin{figure}[t]
	\centering
	\begin{overpic}[height=6.5cm, width=6.5cm]{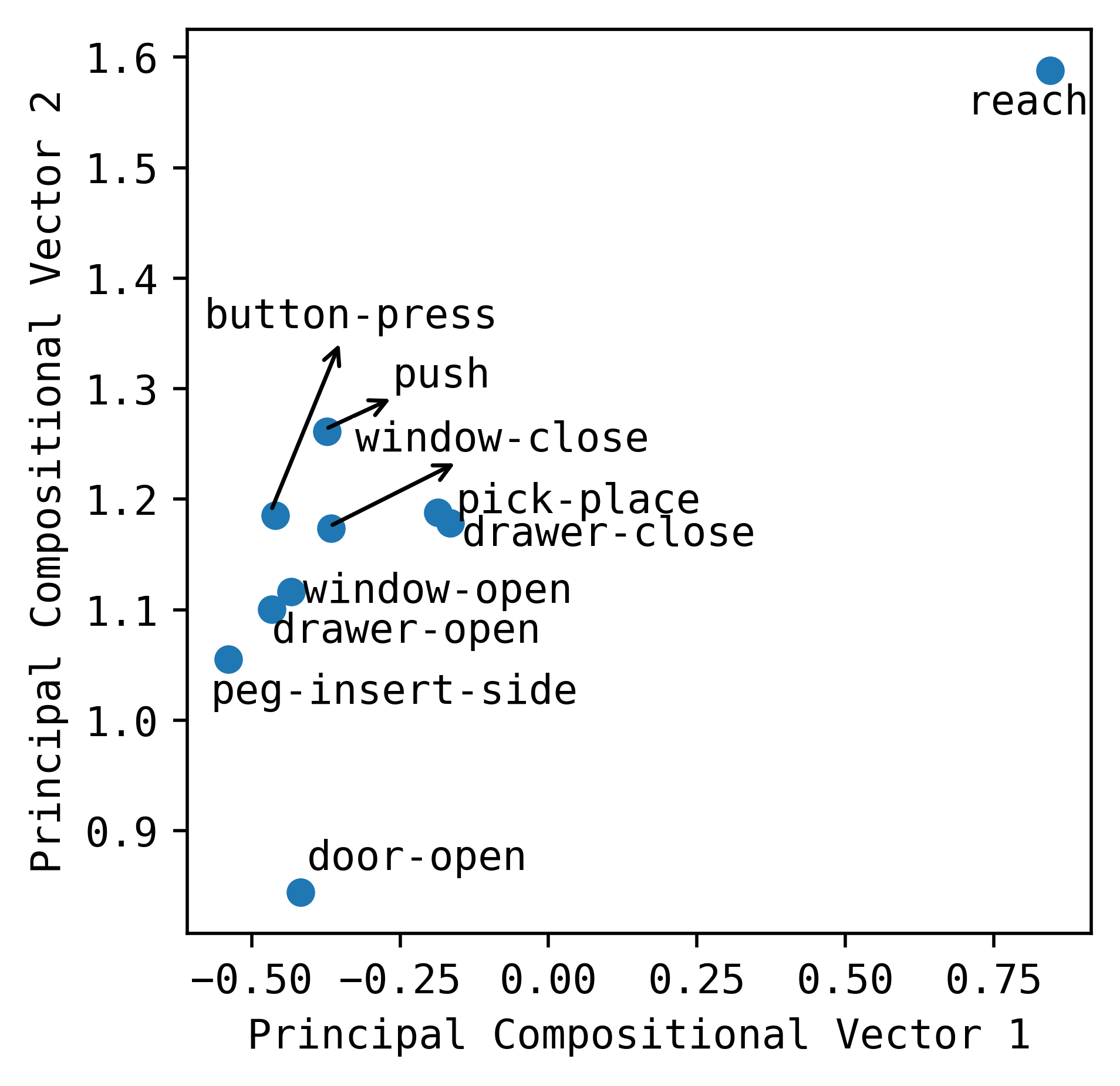}
	\end{overpic}
	\vspace{-0.1in}
	\caption{\rev{PCA projections of $\{\w_\tau\}$ learned using PaCo~\cite{paco} on MT10~\cite{metaworld} benchmark. Each dot represents a projected $\w_\tau$ vector of a task.  Arrows are used to connect the cluttered dots with their respective task names.}}
	\vspace{-0.2in}
	\label{fig:paco_pca}
\end{figure}

\vspace{-0.05in}
\subsection{Connection with Meta Reinforcement Learning}
\vspace{-0.05in}
Meta Reinforcement Learning is one set of approaches for policy reusing and transferring in RL.
Meta-learning or few-shot learning focuses on the fast adaptation of the model parameters with a few data points from the new task.
One commonly used type of gradient-based formulation for meta-learning is a bi-level optimization form as in MAML~\cite{maml}, with the inner loop optimizing for the adapted parameters based on few-shot samples and the outer loop optimizing task loss given the adapted parameters.
Recent progress shows that feature reuse is the main contributing factor in MAML and its bi-level can be reduced to an
almost no inner loop (ANIL) version~\cite{ANIL, bridge_mtl_meta}.
It is important to note that while connected, gradient-based meta-learning~\cite{bridge_mtl_meta, ANIL, maml} mainly focuses on fast adaptation from few-shot data.
\rev{This scheme is effective when there are limited
differences between source and target tasks (\emph{e.g. reach different goal locations, move in different directions, etc.}).
When applied to typical robotics tasks, where the difference between tasks could be large, the few-shot regime is typically not enough and has low
performance in practice (\emph{e.g.} the average success rate over transferred task is relatively low on MetaWorld benchmark~\cite{metaworld}).}
We take an attempt towards solving challenging scenarios involving the transferring beyond simple variations such as goal locations, but between different skills as typical in real-world robotics, thus operating in a zone that is complementary to typical meta RL methods.



\begin{figure*}[t]
	\centering
	\begin{overpic}[width=14cm]{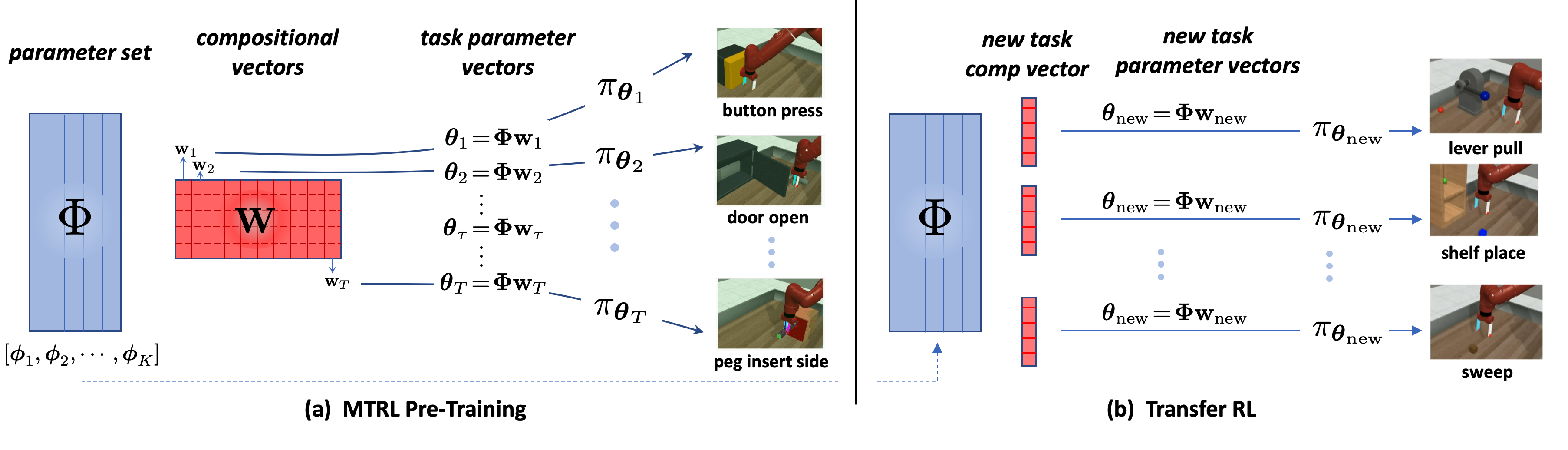}
	\end{overpic}
	\vspace{-0.1in}
	\caption{Transfer with Parameter Compositional MTRL (TaCo). There are two phases in TaCo.
	(a) MTRL pre-training on a set of source tasks. (b) Transfer RL on a set of new tasks with the parameter set learned in the pre-training phase. Here we illustrate with transferring to each new task separately. Transferring jointly to a set of new tasks is also possible with TaCo.}
	\label{fig:framework}
		\vspace{-0.1in}
\end{figure*}


\rev{\section{Review of Parameter Compositional Multi-Task RL (PaCo)}}
\label{sec:paco}

We briefly review the PaCo framework~\cite{paco}, which is also visually depicted in Figure~\ref{fig:framework} (a).
Given a task $\tau\!\sim\!\mathcal{T}$, we use $\vtheta_{\tau} \in \mathbb{R}^{n}$ to denote the vector
of all the trainable parameters for task $\tau$.
The following decomposition for $\vtheta_{\tau}$ is used:
\begin{equation}\label{eq:paco}
\vtheta_{\tau}=\vPhi\w_{\tau},
\end{equation}
where  $\vPhi=[\vphi_1, \cdots, \vphi_i, \cdots, \vphi_K] \in \mathbb{R}^{n\times K}$ denotes a matrix formed by a set of $K$ parameter vectors $\{\vphi_i\}_{i=1}^K$ (referred to as \emph{parameter set}), each of which has the same dimensionality as $\vtheta_{\tau}$, \emph{i.e.}, $\vphi_i \in \mathbb{R}^{n}$.
$\w_{\tau} \!\in\! \mathbb{R}^{K}$ is a \emph{compositional vector}, which is implemented as a trainable embedding for the task index $\tau$.
In essence,  Eqn.(\ref{eq:paco}) decomposes the parameters to two parts: \emph{i)} task-agnostic $\vPhi$ and \emph{ii)} task-aware $\w_{\tau}$.

When faced with multiple tasks as in MTRL, the decomposition in Eqn.(\ref{eq:paco}) offers opportunities for flexible parameter sharing between tasks,
by sharing the task-agnostic $\vPhi$ across all the tasks, while still ensuring task awareness via $\w_{\tau}$, leading to:
\begin{eqnarray}\label{eq:paco_multi}
\begin{split}
[\vtheta_{1}, \cdots, \vtheta_{\tau}, \cdots, \vtheta_{T}] &= \vPhi [\w_{1}, \cdots,  \w_{\tau}, \cdots\w_{T}] \\
\vTheta &=\vPhi\W.
\end{split}
\end{eqnarray}
It also enables a way to stabilize  MTRL training by masking out the exploding loss (when the loss for a particular task $\eta$ is larger than a threshold $\epsilon$, \emph{i.e.} $\mathcal{L}_{\eta} \!>\! \epsilon$) and re-initialize $\w_{\eta}$ as~\cite{paco}:
\begin{equation} \label{eq:reset}
\w_{\eta} =  \sum_{j\in \mathcal{V}} \beta_j \w_j, \quad \vbeta \!=\! [\beta_1, \beta_2, \cdots] \! \sim \! \Delta^{|\mathcal{V}|-1}
\end{equation}
where  $\mathcal{V} \!\triangleq\! \{j| \mathcal{L}_{j} \le \epsilon \}$, and  $\vbeta$ is uniformly sampled from  a unit $|\mathcal{V}|\!-\!1$-simplex $\Delta^{|\mathcal{V}|-1}$.
Our implementation is based on SAC~\cite{sac}.

\section{TaCo: Transfer RL with Improved PaCo}
\label{sec:taco_mtrl}

\subsection{Improved MTRL with Non-Uniform Task Sampling}
During the MTRL stage, we sample tasks for training according to a certain distribution $P$ for T tasks.
There are many potential choices for $P$.
A uniform distribution ($p_{\tau}=1/T$) is typically used in standard MTRL~\cite{soft_module, care, paco}.
A potential way to further improve the MTRL performance is by incorporating non-uniform task sampling during learning using prior or posterior knowledge of training task relationships. \rev{We group tasks into $G$, where $G=\{g_1, g_2, ...g_p\}$, where $g_i$ is a set of tasks, and $\sum_{i=1}^{p}|g_p|=T$. Then we uniformly distribute sample steps among groups (for task $\tau$ in group $g_i$, $p_{\tau}=\frac{1}{|G||g_i|}$)}.

\rev{When we don't have access to pre-defined task groups based on difficulty or similarity before training, we can perform online adjustments on the task distribution (grouping) using the feedback during training(e.g., task relation in Figure~\ref{fig:paco_pca}). A simple approach to doing online adjustment is to perform clustering on task-dependent policy parameters $\theta_{\tau}$. In experiments on \rev{MT10}, we use DBSCAN~\cite{dbscan} to cluster the policy parameters of 10 tasks into groups $G=\{g_1, g_2, ...g_p\}$ and balance the distribution of task groups (referred to as TaCo-online).
}

Alternatively, when information on task groups is available, we can set the distribution of different task groups directly according to this information when desirable.

After a base parameter $\vPhi^*$ is obtained after MTRL training, it can be further used for learning new tasks as described in the next section.



\subsection{Leveraging Previously Learned Shared Parameter Set for Transfer RL}

Following the paradigm of using supervised multi-task learning for transferring~\cite{bridge_mtl_meta}, we train a transferable model via MTRL and then transfer the model to the learning on the target task.
More concretely, we formulate the learning of a transferable model as follows:
\begin{equation}\label{eq:MTRL}
 \text{\small (Phase 1 MTRL)} \;    \vPhi^*, \{\w_{\tau}^*\} = \arg \min_{\vPhi, \{\w_{\tau}\}} \sum_{\tau \in \mathcal{T}}\mathcal{L}(\vPhi, \w_{\tau})
\end{equation}
where $\mathcal{L}$ denotes the standard RL loss~\cite{sac}.
$\mathcal{T}$ denotes the set of tasks with $|\mathcal{T}|\!=\!T$.
After training on a set of tasks $\mathcal{T}$, we obtain a base parameter $\vPhi^*$, where the superscript $^*$ is used to denote the optimal parameter.

While Eqn. (\ref{eq:MTRL}) is general and is compatible with potentially many variants of MTRL approaches,
in this work, we use the Parameter Compositional (PaCo) MTRL approach~\cite{paco} due to its compelling performance and its suitable structure.

\rev{Upon learning of the new task, it is possible to reuse $\vPhi^*$ (and possibly fine-tune from it) and learn a $\w_{\rm new}$ from scratch}.
To achieve this goal, we have
\begin{equation}\label{eq:transfer}
  \text{\small (Phase 2 TransferRL)}   \min_{\vPhi, \w_{\rm new}} \mathcal{L}_{\rm new}(\vPhi, \w_{\rm new}), \, \vPhi \xleftarrow{\text{init}} \vPhi^*
\end{equation}
This scheme of transferring is a path naturally induced from the formulation in Eqn.(\ref{eq:MTRL}).
where $\mathcal{L}_{\vPhi, \w_{\rm new}}$  denote the loss for task $\tau$, $\vPhi$ is the base parameter shared across all tasks, and $\w_{\rm new}$ is the parameter what will be adapted when transferred to new tasks.
Compared with a standard MTRL formulation of $ \min_{\theta} \sum_{\tau \in \mathcal{T}}\mathcal{L}(\theta)$,
Eqn.(\ref{eq:MTRL}) is different in that it explicitly separates all the parameters into two parts.
This separation in parameters could have two implications:
\textbf{\emph{i)}} it facilitates a natural path to transferring (\emph{c.f.} Eqn.(\ref{eq:transfer}));
\textbf{\emph{ii)}} \rev{it introduces a perspective} on the architecture design of the model.
More concretely, it suggests that there should be two types of parameters; one serves for the purpose of retaining previously learned knowledge and can be reused as fixed or fine-tuned parameters for transferring, while the other is more specific that can be learned for each task, including the new task to be transferred to, for which purpose
PaCo fits well.
The transfer RL leveraging the  parameter compositional form  is shown in Figure \ref{fig:framework} (b).


In the transfer RL stage, we reuse pre-trained parameters from the previous MTRL stage. The task-agnostic parameter $\vPhi^*$ consists of both policies value function networks for pre-trained multi-tasks. Using the pre-trained sub-policy space in training can make use of the task similarities between trained and transferred tasks and potentially improve the efficiency in exploration. However, for value functions, reusing the pre-trained parameters usually results in failure in RL training~\cite{simple_approach}. Therefore, during transfer, we only transfer the policy-related parameters  from $\vPhi ^*$ (denoted as $\vPhi \xleftarrow{{\pi}} \vPhi^*$) as initialization and retrain the value function of the new task.

{\flushleft\textbf{Exploration from Transfer Stage.}} The original SAC algorithm starts with collecting data with a random policy. To make full use of the transferred policy subspace $\vPhi$, we replace the policy with random sub-space policy $\pi_{\rm explore} \triangleq \pi_{\vtheta}$, where $\vtheta=\vPhi\tilde{\w}$ and $\tilde{\w} = \W\vbeta$, $\vbeta  \! \sim \! \Delta^K$ and collect $n_e$ steps of data. The value function for the new task is trained during this stage, but the policy parameters remain the same and are released after this stage.

We refer to our instantiation of the general transferring setting Eqn.(\ref{eq:MTRL})$\sim$(\ref{eq:transfer}) with the parameter compositional form Eqn.(\ref{eq:paco})$\sim$(\ref{eq:paco_multi}) incorporating improved training as Transfer with Parameter Compositional MTRL (TaCo).
Detailed algorithmic procedures are presented in the Algorithms below.

\begin{table}[h]
	\vspace{-0.1in}
	\begin{minipage}{0.49\textwidth}
            \begin{algorithm}[H]
            	\centering
            	\caption{TaCo: MTRL Pre-Training Phase}
            	\label{alg:algo}
            			\begin{algorithmic}
            				\STATE {\bfseries Input:}  parameter-set size $K$, loss threshold $\epsilon$, learning rate $\lambda$, task distribution~$P$  \\
            				\WHILE{termination condition not satisfied}
            				\STATE \rev{\emph{\# environmental interaction}}
            				\STATE \rev{sample a task from $P$ upon environment reset and unroll it with the current policy; save task-id and transitions to a replay buffer}\\
            				\STATE \emph{\# policy training}
        					\FOR{\rev{each gradient step}}     
	            				\STATE \rev{sample a batch of training tasks $\mathcal{T}$ and associated data from buffer}
                				\STATE   ${\mathcal{L}}_{\tau} \leftarrow \mathcal{L}_{\tau}(\vtheta_{\tau})$,\, $\vtheta_{\tau} = \vPhi \w_{\tau}$ \;   $\forall \tau \in \mathcal{T}$\\
                				\STATE 	(\emph{loss maskout})  \, $\mathcal{L}_{\eta} \leftarrow  0 \quad \text{if} \; \mathcal{L}_{\eta} > \epsilon$ \\
                				\STATE $\mathcal{L}_{\vTheta}  \leftarrow \sum_{\tau} {\mathcal{L}}_{\tau}$
                				\STATE $\vPhi \leftarrow \vPhi - \lambda \nabla_{\vPhi} \mathcal{L}_{\vTheta} $
                				\STATE $\w_{\tau} \!\leftarrow\! \w_{\tau} \!-\! \lambda \nabla_{\w_{\tau}} \mathcal{L}_{\tau}(\w_{\tau}) $
                				\STATE 	($\w$-reset) \, $\w_{\eta} \leftarrow  \text{Eqn.(\ref{eq:reset})} \quad \text{if} \, \mathcal{L}_{\eta} > \epsilon\,$
        					\ENDFOR
            				\ENDWHILE
            			\end{algorithmic}
            \end{algorithm}
	\end{minipage}
	\hspace{0.1in}
	\begin{minipage}{0.465\textwidth}
            \begin{algorithm}[H]
            	\centering
            	\caption{TaCo: Transfer RL Phase}
            	\label{alg:algo2}
            			\begin{algorithmic}
            				\STATE {\bfseries Input:} MTRL  parameter set $\vPhi^*$
            				\STATE Initialize $\vPhi \xleftarrow{{\pi}} \vPhi^*$
            				\STATE Collect $n_e$ steps of data with $\pi_{\rm explore}$, update value functions.
            				\WHILE{termination condition not satisfied}
            				\STATE   $\vtheta_{\rm new} = \vPhi \w_{\rm new}$  \\
            				\STATE ${\mathcal{L}}_{\rm new} \leftarrow \mathcal{L}_{\rm new}(\vtheta_{\rm new})$
            				\STATE $\vPhi \leftarrow \vPhi - \lambda \nabla_{\vPhi} \mathcal{L}_{\rm new} $
            				\STATE $\w_{\rm new} \!\leftarrow\! \w_{\rm new} \!-\! \lambda \nabla_{\w_{\rm new}} \mathcal{L}_{\rm new} $
            				\ENDWHILE
            			\end{algorithmic}
            \end{algorithm}
	\end{minipage}
    \\
	\vspace{-0.1in}
\end{table}

\section{Experiments}
\label{sec:result}
In this section, we empirically test the performance of Multi-task RL and transfer RL on the Meta-World benchmark~\cite{metaworld}. Meta-World benchmark is a robotic environment consisting of a number of distinct manipulation tasks, with example tasks shown in Figure \ref{fig:example_tasks}. All the tasks share the same action space and the same dimension of state space, but certain dimensions in the state space represent different semantic meanings across tasks. 
We used the Multi-task benchmark from Meta-World~\cite{metaworld} with random goals, i.e., each manipulation task is configured with random goals, and the 10-task benchmark is referred to as \rev{MT10}.

Using experiments on Meta-World, we would like to demonstrate: \textbf{\emph{i)}} TaCo's state-of-the-art performance for MTRL,  and \textbf{\emph{ii)}} TaCo's advantages and effectiveness in transfer learning on unseen tasks using a pre-trained multi-task policy.
To make the presentation clear, we present results on each stage in  subsection~\ref{exp:MTRL} and \ref{exp:Transfer} respectively.

\subsection{MTRL Experiments and Results}\label{exp:MTRL}
{\flushleft\textbf{Training Setting.}}
For MTRL training on \rev{MT10} in Meta-World, we follow the settings introduced in \cite{care} and  use \textbf{\emph{i)}} 10 parallel environments, \textbf{\emph{ii)}}~20 million environment steps for the 10 tasks together (2 million per task), \textbf{\emph{iii)}}~repeated training with 10 different random seeds for each method.

{\flushleft{\textbf{Baselines.}}}
For MTRL benchmarks in Meta-World, we compare it against \emph{(i)} \textbf{Multi-task SAC}: extended SAC~\cite{sac} for MTRL with one-hot task encoding;  \emph{(ii)} \textbf{Multi-Head SAC}: SAC with shared a network apart from the output heads, which are independent for each task; \emph{(iii)}  \textbf{SAC+FiLM}: the task-conditional policy is implemented with the FiLM module~\cite{film} on top of SAC; \emph{(iv)} \textbf{PCGrad}~\cite{pcgrad}: a representative method for handling conflicting gradients during multi-task learning via gradient projection during optimization; \emph{(v)} \textbf{Soft-Module}~\cite{soft_module}: which learns a routing network that guides the soft combination of modules (activations) for each task; \emph{(vi)} \textbf{CARE} \cite{care}: leveraging additional task-relevant metadata for state representation. \emph{(vii)} \textbf{PaCo}~\cite{paco}: Parameter-Compositional MTRL with uniform task distribution.\footnote{Note that some experiments reported in the works mentioned above are implemented and evaluated on Meta-World-V1 and/or with fixed-goal setting. We adapt these methods and experiment on Meta-World-V2.}

\begin{table}
   \caption{Results on Meta-World~\cite{metaworld} \rev{MT10} (20M steps).}
    \centering
	\resizebox{6.cm}{!}{
    \begin{tabular}{l|c}
			\hline
			\multicolumn{1}{c|}{\multirow{2}{*}{Methods}} & Success Rate ($\smallpercent$) \\
			& (mean $\pm$ std) \\
			\hline \hline
			Multi-Task SAC~\cite{metaworld} & \multicolumn{1}{l}{\quad 62.9 $\pm$ 8.0} \\
			Multi-Head SAC~\cite{metaworld} & \multicolumn{1}{l}{\quad 62.0 $\pm$ 8.2} \\
			SAC + FiLM~\cite{film} & \multicolumn{1}{l}{\quad 58.3 $\pm$ 4.3} \\
			PCGrad~\cite{pcgrad} & \multicolumn{1}{l}{\quad 61.7 $\pm$ 10.9} \\
			Soft-Module~\cite{soft_module} &\multicolumn{1}{l}{\quad  63.0 $\pm$ 4.2} \\
			CARE~\cite{care} & \multicolumn{1}{l}{\quad 76.0 $\pm$ 6.9}   \\
			PaCo~\cite{paco} & \multicolumn{1}{l}{\quad 85.4 $\pm$ 4.5}\\
			\hline
    		\multicolumn{1}{l|}{\textbf{TaCo} (Ours)} & \multicolumn{1}{l}{\quad \textbf{90.7 $\pm$ 3.6}}\\
			\hline
		\end{tabular}
		}
	\label{tab:MT10}
\end{table}

{\flushleft\textbf{Evaluation Metrics and Results.}}
The evaluation metric for the MTRL policy is based on the success rate of the policy for all the tasks. For Meta-World benchmarks, we evaluate each skill with 5 episodes of different sampled goals using the final policy. The success rate is then averaged across all the skills. There is randomness in the MTRL training; therefore instead of picking the maximum evaluation success rate across training, we use the policy at 20M total environment steps (2M per task) and average across multiple trains for a fair evaluation. 
\rev{The 20M total environmental steps (i.e. 2M per task) metric is chosen based on previous work~\cite{care, paco} which empirically observed that all the methods have mostly saturated in their performance under this setting.}
We report the mean and standard deviation of the success rate in Table \ref{tab:MT10}. An improvement on the \rev{MT10} benchmark is observed on PaCo~\cite{paco} compared to previous MTRL methods, demonstrating the advantage of the compositional structure and the reset during training. A further improvement of TaCo using pre-set task distribution during the MTRL training is also observed, showing the effectiveness of using task difficulty to adjust the task distribution of different tasks.

{\flushleft\textbf{Adjustment of Task Distribution.}}
An important component that contributes to the significant improvement of TaCo on MTRL is the adjusted sample distribution according to task difficulties, as discussed in Section~\ref{sec:taco_mtrl}. 
We investigate the impact of this factor empirically in this section.
The comparison between PaCo (uniform task distribution) and TaCo-online is shown in Table~\ref{tab:MT10_abl_task_distribution}. 
We evaluate the performance  at 20M step and also provide an evaluation at 30M steps to further inspect the potential performance changes w.r.t. increased environmental steps. 
As expected, the TaCo-online adjustment using task grouping has an improvement over the PaCo baseline.
When additional information is available for designing task distributions, TaCo can further improve its performance.
To demonstrate this, we empirically construct a task distribution as follows.

\begin{figure}
  \begin{center}
    \includegraphics[width=7cm]{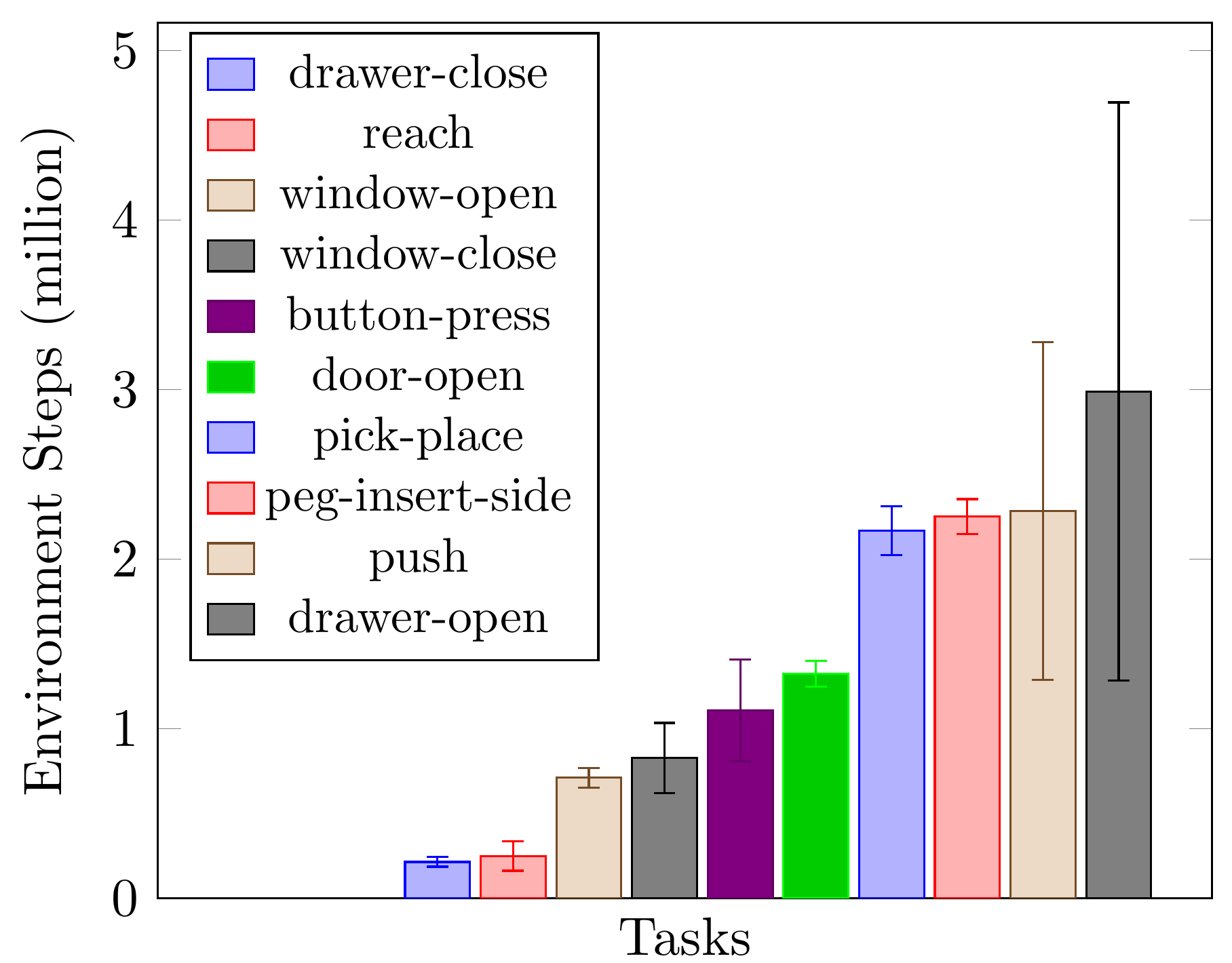}
  \end{center}
  \vspace{-0.25in}
	\caption{Environment steps required for training a successful single-task policy for different tasks.}
	\label{fig:env_steps}
	\vspace{-0.25in}
\end{figure}

\begin{table}[h]
    \caption{Different task distributions in MTRL training (\rev{MT10} Tasks).}
    \centering
	\resizebox{6.5cm}{!}{
    \begin{tabular}{l|c|c}
			\hline
			\multicolumn{1}{c|}{\multirow{2}{*}{Methods}} & \multicolumn{2}{c}{Success Rate ($\smallpercent$)} \\
			& \rev{20M} steps & 30M steps \\
			\hline \hline
			PaCo~\cite{paco} & \multicolumn{1}{l|}{\quad \rev{85.4 $\pm$ 4.5}} & \multicolumn{1}{l}{\quad 85.7 $\pm$ 4.1} \\
			\hline
			TaCo-online  &  \multicolumn{1}{l|}{\quad \rev{86.4 $\pm$ 4.3}}  & \multicolumn{1}{l}{\quad 89.3 $\pm$ 1.2} \\
			\textbf{TaCo}   & \multicolumn{1}{l|}{\quad \textbf{\rev{90.7 $\pm$ 3.6}}}     & \multicolumn{1}{l}{\quad \textbf{94.0 $\pm$ 5.4}} \\
			\hline
		\end{tabular}
		}
	\label{tab:MT10_abl_task_distribution}
	\vspace{-0.1in}
\end{table}
We set the probability of ``\textit{pick-place, peg-insert-side, drawer-open}'' two times the probability of other tasks, based on the descriptions of each task in \rev{MT10}.\footnote{The ``difficulty'' of task is based on our intuition on reward sparsity based on task description, we do not assume single task training statistics as prior knowledge like~\cite{simple_approach}.} As shown in Table~\ref{tab:MT10_abl_task_distribution}, TaCo achieves higher success rate with this task distribution.
As a reference, the environment steps required for single tasks to converge are shown in Figure \ref{fig:env_steps}.

\subsection{Transfer Experiments and Results}
\label{exp:Transfer}
{\flushleft\textbf{Training Setting.}} For transfer learning with MTRL policy, we use the best policy (in terms of average success rate) during MTRL training as the base policy. We use 10 parallel environments and 5 random seeds for each transfer experiment. During the transfer, we set the initial  exploration steps for data collection as $n_{\rm warm}\!=\!20$K.

\begin{figure*}[h]
	\centering
	\begin{overpic}[width=14.cm]{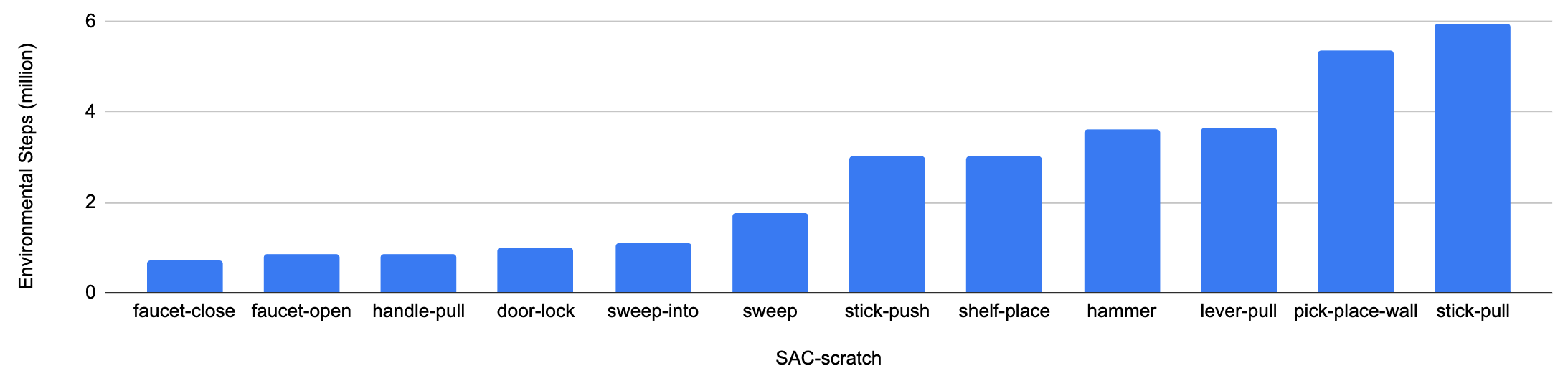}
	\end{overpic}
	\begin{overpic}[width=14.cm]{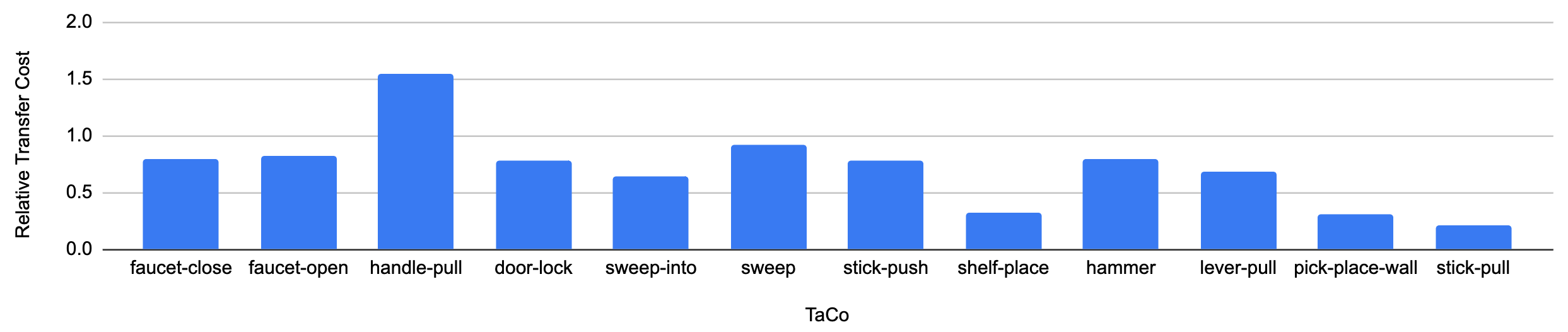}
	\end{overpic}
	\vspace{-0.15in}
	\caption{More transfer results. (a) The number of environmental steps required for SAC-scratch for 12 different tasks. \rev{Note that only successfully trained seeds are used.}
	(b)~Relative Transfer Cost of TaCo over SAC-scratch. \rev{Training success rate and step cost for SAC and TaCo are both considered}.}
	\label{fig:12task_results}
	\vspace{-0.25in}
\end{figure*}

\vspace{-0.05in}
{\flushleft\textbf{Evaluation Metrics.}}
From Figure~\ref{fig:env_steps}, we observe that the \rev{MT10} benchmark actually covers a wide range of tasks in terms of single-task difficulty. These required environment steps can be used as calibration.
The following metrics are used in reporting the results:
\begin{enumerate}[leftmargin=*]
    \item \textbf{required environmental steps} $n$: the number of environmental steps (M, in million) required to reach an evaluation success rate of 0.9, before $n_{\rm max}$ steps ($n_{\rm max}=6\rm M$ steps in evaluation)
    \item \textbf{(transfer) success} $\alpha$: for every single task, we train for 6M environmental steps. It is labeled as a success if the training reaches an evaluation success rate of 0.9.
    \item \textbf{relative transfer cost} (main metric): ratio of the success-normalized environment steps ($\frac{n}{\alpha}$) used to train a new task with and without the transferring process: $\frac{n_{\rm transfer}}{\alpha_{\rm transfer}} / \frac{n_{\rm scratch}}{\alpha_{\rm scratch}}$.
\end{enumerate}
We repeat the training 5 times with different random seeds and use the average as its empirical estimation for each metric. Results are summarized  in Figure~\ref{fig:12task_results}.


\vspace{-0.05in}
{\flushleft\textbf{Illustrative Transferring within \rev{MT10} Tasks.}}
To get an illustration of how multi-task policies can help train new tasks more efficiently, we first design some transfer learning experiments within the \rev{MT10} tasks. In Table \ref{tab:transfer_3to1}, we show the transfer performance of pre-training on three tasks and transfer to a new task in \rev{MT10}. In general, transfer from MT policy can benefit training on new tasks, the improvement can be relatively large for harder tasks if diverse skills are involved in pre-training.
At the same time, there are cases where the new tasks are simple and unrelated to the pre-trained tasks. The transfer process may cost more steps compared to training from scratch. This phenomenon matches the observation on the transfer metric under a different setting (transferring between single tasks)  reported in~\cite{simple_approach}, where transfer cost  between individual tasks could sometimes increase to a \rev{value larger than 1 depending on} the task relationships.

\begin{table*}[h]
    \caption{Transfer cost and success rate between selected tasks in \rev{MT10}.}
    \vspace{-0.1in}
    \centering
	\resizebox{12.5cm}{!}{
    \begin{tabular}{wc{5cm}|wc{2cm}|c|wc{1.5cm}}
        \hline
        \multirow{2}{*}{Trained Tasks} &  \multirow{2}{*}{New Task} & \multicolumn{2}{c}{Transfer Cost ($\downarrow$) / Trans. Success ($\uparrow$)}\\
        \cline{3-4}
        & & SAC-scratch~\cite{sac} & TaCo \\
        \hline \hline
        \emph{reach, door-open, drawer-open} &  \emph{drawer-close} & 1.0/1.0 & \textbf{0.727/1.0} \\
        \emph{reach, door-open, drawer-open} &  \emph{window-close} & 1.0/1.0 & \textbf{0.879/1.0} \\
        \emph{reach, door-open, drawer-open} &  \emph{window-open}  & \textbf{1.0/1.0} & 1.126/1.0 \\
        \emph{reach, push, peg-insert}       &  \emph{door-open}    & 1.0/1.0 & \textbf{0.528/1.0}\\
        \hline
    \end{tabular}
    }
    \label{tab:transfer_3to1}
    \vspace{-0.1in}
\end{table*}

{\flushleft\textbf{Transfer to New Tasks using MT Policy from \rev{MT10}.}}
Compared to arbitrarily selected task combinations, the \rev{MT10} benchmark covers a set of different skills in manipulation with different difficulties. We now test its performance on transferring to some challenging unseen new tasks.

\rev{We chose four tasks (\textit{sweep}, \textit{sweep-into}, \textit{lever-pull}, and \textit{shelf-place}) from Meta-World that are not
in \rev{MT10}, but are representative of the rest of Meta-World and
experience poor performance during few-shot transfer in reported
results~\cite{metaworld})}.
Training from scratch for these tasks all require more than 1 million steps, which means they are not easy tasks.\footnote{Task order from simple to hard based on single task training success rate and required environment steps is: \textit{sweep-into}, \textit{sweep}, \textit{lever-pull}, \textit{shelf-place}. }
Besides \textbf{SAC-scratch}~\cite{sac}, we further compare with a recent transferring approach called \textbf{ANIL}~\cite{ANIL}, which is an \emph{Almost No Inner Loop} version of MAML~\cite{maml}. It learns the MTRL policy with a multi-head form and then adapts the output layer for transferring to the new task. For reference, it achieves a $70\%$ success rate at convergence on the trained tasks on \rev{MT10}.

In Table~\ref{tab:transfer_MT10}, we show the transfer cost and the transfer success of transferring from a fully trained (reach $100\%$ success for all \rev{MT10} tasks) TaCo MTRL policy.
For TaCo transfer, we observe an improvement in both transfer cost and transfer success. For tasks that occasionally fail in training from scratch, transferring with TaCo improves their stability.
ANIL~\cite{ANIL} has a larger transfer cost compared to from scratch on some tasks, and cannot always match the success rate of training from scratch. This is potentially due to its limited performance during the MTRL stage on \rev{MT10}, and also the large deviation between trained and new tasks compared to typical meta-learning settings. This emphasizes the importance of using a performant MTRL method and the design of an
architecture that is effective for handling large task derivations for effective transferring.

\begin{table*}[h]
    \centering
    \caption{Transfer cost and success rate from \rev{MT10} TaCo policy to new tasks.}
    \vspace{-0.1in}
	\resizebox{13.8cm}{!}{
    \begin{tabular}{wc{5.5cm}|wc{1.8cm}|c|wc{2.0cm}|wc{1.5cm}}
        \hline
\multirow{2}{*}{Trained Tasks} &  \multirow{2}{*}{New Task} & \multicolumn{3}{c}{Transfer Cost ($\downarrow$) / Trans. Success($\uparrow$)}\\
        \cline{3-5}
        && SAC-scratch~\cite{sac} & ANIL~\cite{ANIL} & TaCo \\
        \hline \hline
        \multirow{4}{6cm}{\rev{MT10} \emph{(
        reach,
        push,
        drawer-open,
        drawer-close,
        window-open,
        window-close,
        button-press,
        door-open,
        pick-place,
        peg-insert-side
        )}} & \emph{sweep}        &  1.0/1.0  & \textbf{0.826/1.0} & 0.922/1.0  \\
        & \emph{shelf-place}      &  1.0/0.4  & 1.868/0.4 & \textbf{0.809/1.0} \\
        & \emph{sweep-into}       &  1.0/1.0  & 1.758/1.0 & \textbf{0.654/1.0}\\
        & \emph{lever-pull}       &  1.0/0.8  & 1.232/0.6 & \textbf{0.849/1.0} \\
        \hline
    \end{tabular}
    }
    \label{tab:transfer_MT10}
\end{table*}


{\flushleft\textbf{Transfer Performance on Hard Tasks}}.
Shelf-place is a hard task that requires a huge number of steps in single-task training and is not stable. Nearly half of the experiments launched for the shelf-place task fail to converge to a successful policy due to a lack of exploration or loss explosion. Therefore, compared to other tasks, where efficiency of learning serves as the main metric for transfer learning, stability improvement, as well as the environment step efficiency, are both important.
\begin{figure}
  \begin{center}
    \includegraphics[height=6.0cm]{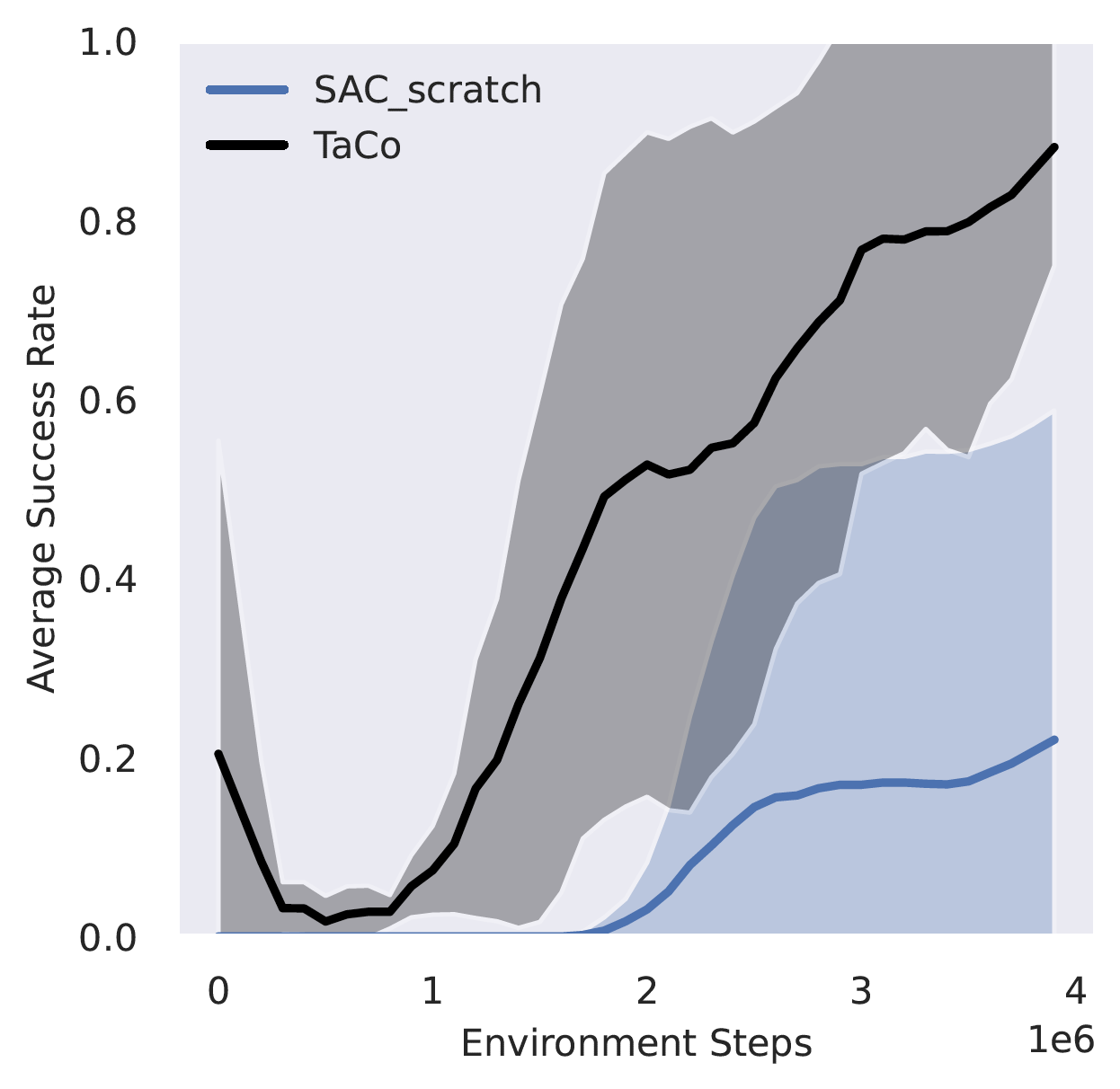}
  \end{center}
  \vspace{-0.3in}
	\caption{Comparison of training from scratch and transfer from TaCo-\rev{MT10} on \textit{shelf-place}, which is a  hard task.}
	\label{fig:shelf_place}
	\vspace{-0.2in}
\end{figure}
Figure \ref{fig:shelf_place} shows the comparison of the average success rate curve between training from the transferred policy and training from scratch.
Note that the improvement we can see from using transferred multi-task policy largely benefits from the good exploration performed by random policies in the trained policy subspace. This is not guaranteed for arbitrary new tasks and multi-task sets. There are chances that the trained multi-task policy doesn't benefit the new training when the new task is fundamentally different from the trained tasks. This is also a non-negligible challenge we have for transfer learning. However, \rev{by introducing more diversity} in the pre-trained task set, it is more likely a new task can benefit from the general multi-task policy.

\vspace{-0.05in}
{\flushleft{\textbf{More Experiments on TaCo Transfer.}}}
We have the following observations from the results in Figure~\ref{fig:12task_results}:
\begin{enumerate}[leftmargin=*]
    \item \textbf{reduced environment steps}: TaCo transfer can reduce the training cost by using fewer environment steps to converge on most tasks, demonstrating the first benefit of transferring;
    \item \textbf{improved success rate}:  TaCo also shows an improved success rate, which is more distinct for hard tasks, where training from scratch has a success rate lower than those of easy tasks, implying the second benefit;\footnote{The relative difficulty of a task can be roughly measured by the number of environment steps required for obtaining a well-performing policy.}
    \item \textbf{reduced relative cost}: in terms of the relative transfer cost defined above, which is a metric that incorporates both the aspect of transfer cost and success rate, TaCo also shows clear improvements compared with SAC-scratch. The reduction is especially clear for difficult tasks (\emph{e.g.} tasks with environmental steps$>$2M according to Figure~\ref{fig:12task_results}(a)). For example, for the \emph{stick-pull} task,
    SAC-scratch spends roughly 6M environment steps with a 0.2 success rate. TaCo on the other hand, increases the success rate to 0.8 with fewer environmental steps, leading to a significant
    reduction in transfer cost.
    \item \textbf{transfer performance variations}: while there are clear improvements in most tasks, we do see \rev{in tasks like \emph{handle-pull}}, training from scratch is already stable in succeeding and is more efficient in environment steps. This is potentially due to its low task difficulty and the lack of task similarity between the \rev{MT10} task set and \emph{handle-pull}.
    How to characterize the similarities between trained and new tasks is another open research question and an interesting topic for future work.
\end{enumerate}

\vspace{-0.05in}
\subsection{Additional Example: Transferring with Fixed $\Phi$}
\vspace{-0.05in}
An interesting attempt in TaCo framework is to verify if we can achieve successful transfer with the \emph{fixed} task-agnostic policy parameters learned from pre-trained multiple tasks. In this setting, we fix the learned $\mathbf{\Phi}$ parameters \rev{and learn only} a new $\mathbf{w}$ vector for the unseen task, meaning the new policy lies in the trained multi-task subspace and uses no additional policy parameters except a new $\mathbf{w}$ parameter representing its position in the subspace.
The results are summarized in Table~\ref{tab:transfer_fixed_phi}.
The performance under this setting depends more heavily on the relationships between trained and extension/new tasks. The relation analysis of tasks and sequencing during training is an open problem for robotic learning, and because of this, we report the effectiveness of TaCo's no-cost extension ability as an additional example, acknowledging the non-negligible efforts required to fully investigate this orthogonal direction in the future.

\vspace{-0.1in}
\begin{table}[h]
    \caption{TaCo Transferring with Fixed $\Phi$.}
    \vspace{-0.1in}
    \centering
    \resizebox{8.5cm}{!}{
    \begin{tabular}{ c|c|c}
        \hline
        Trained Tasks &  Extension Task &  Success Rate\\
        \hline
        \emph{reach, door-open, drawer-open} &  \emph{drawer-close}  & $75\!\pm\!9$ \\
        \emph{window-open, window-close, door-open} & \emph{door-close} & $90\!\pm\!5$\\
        \hline
    \end{tabular}
    }
    \vspace{-0.2in}
    \label{tab:transfer_fixed_phi}
\end{table}

\section{Conclusions and Future Work}
\vspace{-0.02in}
\label{sec:conclusion}
In this work, we present TaCo, an approach that leverages MTRL for transfer reinforcement learning.
With TaCo, we can achieve better MTRL performance and effective transfer learning of new tasks on top of it.
We show encouraging results along the direction of leveraging multi-task learning for transferring~\cite{bridge_mtl_meta} under a number of different settings in the RL domain.
Successful transfer hinges on the relationships between the source tasks and the new task.
How to quantify the  diversity of the source tasks and the relationship with the new task is a challenging and interesting future work.

In this work, we have used task id to represent different tasks following protocols in standard MTRL environments~\cite{metaworld}.
\rev{While this setting has minimal assumptions on the tasks, it is limited by its power in encoding task properties and their mutual relationships.
Moreover, it incurs additional labeling costs.
It is possible to improve by using language-based task description~\cite{care} or unsupervised task relationship discovery.}
Another limitation is the evaluation protocol. To the best of our knowledge, there is no standardized evaluation protocols available for MTRL-based transferring in the community, including the selection of the base trained task and new task.
In this work, we used an existing MTRL benchmark with a reasonable diversity, with the hope to retain some relevance to the new task.  How to achieve this in a more principled way requires further investigation.



\rev{While this work mainly focuses on the algorithmic aspects of pre-training and transferring schemes, it is important to note that
transfer learning is a very challenging task that involves many other aspects, including task relations, imbalances, and long-term dependencies. Multi-task and transfer learning in task space needs to combine with sim-to-real techniques considering vision representation \cite{sun_grasping} and task-dependent constraints \cite{wang2023simple} to work on real robots.}
We leave the investigations on these aspects as future work.

\vspace{-0.1in}
\begin{table}[h]
        \caption{General MTRL hyper-parameters on \rev{MT10}}
        \vspace{-0.15in}
    	\centering
		\tabcolsep=0.3cm
		\centering
		\resizebox{6cm}{!}{
		\begin{tabular}{l| c}
			\hline
			Hyper-parameter & Value \\
			\hline \hline
			batch size &  1280 \\
			number of parallel env & 10 \\
			MLP hidden layer size & [400, 400, 400] \\
			policy learning rate & 3e-4\\
			Q learning rate & 3e-4\\
			discount & 0.99\\
			episode length & 150\\
			exploration steps & 1500\\
			replay buffer size & 1e6 \\
			\hline

			\hline
		\end{tabular}
			}
			\vspace{-0.25in}
    \label{tab:general_hyperparameter}
\end{table}

\begin{table}[h]
    \caption{TaCo specific hyper-parameters}
    \vspace{-0.1in}
    \centering
		\tabcolsep=0.3cm
		\resizebox{7cm}{!}{
		\begin{tabular}{p{4.5cm}| c}
    			\hline
    			Hyper-parameter & Value \\
    			\hline \hline
    			extreme loss threshold $\epsilon$ &  3e3 \\
    			param-set size $K$ & 5\\
    			compositional vector $w$ learning rate & 3e-4 \\
    			transfer stage exploration step $n_e$ & 20000 \\

    			\hline

    			\hline
    		\end{tabular}
			}
			\vspace{-0.1in}
    \label{tab:paco_hyperparameter}
\end{table}







\bibliographystyle{IEEEtran}
\bibliography{paco_transfer}

\end{document}